\documentclass[runningheads]{llncs}
\usepackage{graphicx}

\usepackage{tikz}
\usepackage{comment}
\usepackage{amsmath,amssymb}
\usepackage{color}
\usepackage{algorithm}
\usepackage{algpseudocode}
\usepackage{url}
\usepackage{booktabs}

\usepackage[accsupp]{axessibility}
\usepackage[utf8]{inputenc}
\usepackage[english]{babel}

\usepackage[width=122mm,left=12mm,paperwidth=146mm,height=193mm,top=12mm,paperheight=217mm]{geometry}
\usepackage{graphicx,subcaption}

\begin{document}
\footnotesize
\title{Soiling detection for Advanced Driver Assistance Systems}
% ZKONTROLOVAT PODEKOVANI

\titlerunning{Soiling Segmentation}

% PORADI AUTORU PROZATIM PSEUDONAHODNE
\author{
Filip Beránek\inst{1}\orcidID{0009-0002-3563-3097} \and
Václav Diviš\inst{1}\orcidID{0000-0001-9935-7824} \and 
Ivan Gruber \inst{1}\orcidID{0000-0003-2333-433X}
}

\authorrunning{Filip Beránek et al.}

\institute{Department of Cybernetics and New Technologies for the Information Society, Technická 8, 301 00 Plzeň, Czech Republic \email{bionothi@ntis.zcu.cz}\\}
%%%%%%%%%%%%%%%%%%%%%%%%%%%%%%%%%%%%%%%%%%%%%%%%%%%%%%%%%%%%%%

\maketitle              % typeset the header of the contribution

\begin{abstract}
\footnotesize
Soiling detection for automotive cameras is a crucial part of advanced driver assistance systems to make them more robust to external conditions like weather, dust, etc. In this paper, we regard the soiling detection as a semantic segmentation problem. We provide a comprehensive comparison of popular segmentation methods and show their superiority in performance while comparing them to tile-level classification approaches. Moreover, we present an extensive analysis of the Woodscape dataset showing that the original dataset contains a data-leakage and imprecise annotations. To address these problems, we create a new data subset, which, despite being much smaller, provides enough information for the segmentation method to reach comparable results in a much shorter time. All our codes and dataset splits are available at~\url{https://github.com/filipberanek/woodscape_revision}.
% Automotive cameras are essential part of Advanced Driver Assistance Systems (ADAS) and they tend to be exposed by external factors like rain, mud, snow, dust, leaf etc. This is reducing reliability of ADAS systems. Valeo therefore published first dataset with soiling annotations, where we can detect where is soiling located and what is the level of occlusion. We enhance their research by comparing their solution to semantic segmentation and by revising published dataset and its quality. We also dive into definition of occlusion level as it is very subjective and must be precisely defined. With this work we want to encourage further research of adverse condition effects. \url{https://github.com/filipberanek/woodscape_revision}

\end{abstract}

\section{Introduction}
In pursuit of safer roadways, automotive manufacturers are using the power of Advanced Driver Assistance Systems (ADAS)~\cite{ADAS_Review}. These systems, ranging from Adaptive Cruise Control (ACC) to Lane-Keeping Assist (LKA), represent a critical step toward fully autonomous vehicles~\cite{Autonomous_cars}. Additionally, comfort-oriented features like parking assistance further improve the driving experience. The central role of the functionality of ADAS is playing various sensors such as lidars, radars, ultrasound, and cameras. While these sensors perform reliably under optimal weather conditions, real-world scenarios present challenges. Factors like occlusion, particularly when sensors are exposed and unshielded, can significantly compromise system reliability.

Until recently, there has been a small number of research addressing poor visibility conditions. This changed with the introduction of the UG2 competition at CVPR 2019~\cite{UG2}. Among others, the Valeo company addressed this research gap by releasing the first dataset featuring occluded camera imagery, known as the Woodscape dataset~\cite{Woodscape}. Alongside the dataset they released extensive work on occlusion detection~\cite{Soilingnet}\cite{TiledSoilingNet}\cite{Soiling_data_augmentation}\cite{Ensemble_based_semi_supervised_soiling_learning}.

Soiling detection extends beyond mere identification. It encompasses evaluating the level of occlusion and location. Consider a scenario where a small portion of an image is heavily obscured, impacting crucial decisions like lane-keeping or collision avoidance. Hence, understanding the position and level of soiling becomes imperative.

Building upon Valeo's pioneering work, we extend their research. Previous articles~\cite{Soilingnet}\cite{TiledSoilingNet} mention semantic segmentation as a potential solution, but only tile-level detection has been explored. In this paper, we expand on the previous research by conducting baseline experiments with semantic segmentation models. Moreover, we provide an indirect comparison to published results by Valeo. 

% Unfortunately, a direct comparison is not possible as Valeo published only a part of the original Woodscape dataset. 

Lastly, we dive into the Woodscape dataset. Starting with the dataset and its split revision, where we would like to explore how the data split is done. Then we explore the impact of annotation quality. Given the subjective nature of soiling annotation, where the same occlusion could be identified by different annotators as different levels of occlusion, achieving precise distinctions between classes poses a challenge, as discussed in~\cite{Soiling_data_augmentation} and~\cite{Ensemble_based_semi_supervised_soiling_learning}. We aim to evaluate annotation quality, present basic dataset statistics, and assess how variations in annotation quality affect semantic segmentation performance.

Our contributions can be summarized as follows:
\begin{itemize}
\item A comprehensive analysis of the Valeo's Woodscape dataset, including dataset split refinement.
\item An introduction of semantic segmentation results and their comparison to tile-level detection.
\item An examination of the influence of annotations refinement on the segmentation accuracy.
\end{itemize}
\section{Related Work}
Research related to occlusion or adverse conditions has three main directions. Firstly, denoising strategies endeavor to eliminate occlusions from camera images. For instance, Porav et al.~\cite{porav2019i} focus on removing water droplets from camera lenses. Other notable works include those by~\cite{single_image_dehazing}\cite{chen2023deanet}\cite{non_local_image_dehazing}\cite{image_dehazing_ultra_high_res}\cite{Denoising}, all dedicated to restoring images to their original clarity by mitigating occlusions.

Secondly, efforts are made to adapt networks in order to work better under adverse conditions. Works such as those by Sakaridis et al.~\cite{Sakaridis_2018}\cite{sakaridis2018model}\cite{pfeuffer2019robust}\cite{alshammari2019multitask} focus on foggy scenes, striving to execute semantic segmentation accurately under challenging conditions.

Thirdly, the detection of occlusions, primarily spearheaded by Valeo, is highlighted in the following works~\cite{Soilingnet}\cite{TiledSoilingNet}\cite{Soiling_data_augmentation}\cite{Ensemble_based_semi_supervised_soiling_learning}.

All these approaches face a common obstacle: the scarcity of data. This situation has been alleviated by initiatives like~\cite{UG2}. Additionally, Valeo Woodscape dataset~\cite{Woodscape}, the first soiling dataset published, has paved the way for further research in this domain.

Most of the previous research focused predominantly on developing new and improving already existing algorithms. However, the most important aspect of supervised learning, the data, was neglected. Introduced by Andrew Ng~\cite{data_centric}, Domain or Data-Centric AI (DCAI) is currently increasing in importance in AI research. There has been also published Six principles of DCAI by Jarrahi et al.~\cite{jarrahi2022principles} and further evaluation of large-scale Automotive datasets, conducted by Divis et al.~\cite{divivs2023domain}. This research highlighted the potential influence of unbalanced features and proposed methods how to evaluate and collect uniformly feature-distributed domain-related datasets.

% In this study, we leverage the Valeo Woodscape dataset and utilize baseline semantic segmentation networks to establish metrics for comparison. These baseline architectures, including Unet \cite{unet}, Unet++ \cite{unetplutplus}, DeepLabV3 \cite{deeplabv3}, DeepLabV3++ \cite{deeplabv3plusplus}, FPNet \cite{fpn}, PSPNet \cite{pspnet}, MaNet \cite{manet}, LinkNet \cite{linknet}, and PAN \cite{pan}, are implemented within the \cite{pytorch_segmentation} framework.
\section{Dataset analysis}
\label{03_dataset_analysis}
The Woodscape dataset initially comprised 76,448 images and their corresponding labels~\cite{Woodscape}\cite{Soilingnet}, however, only a fraction of this dataset has been made publicly available, totaling 5,000 images with the labels. The dataset is divided into two subsets: the "Train set," containing 4,000 images, and the "Test set", which contains 1,000 images. Notably, the images were captured sequentially, typically featuring 7 or 8 consecutive frames from the same scene and camera setup. In the dataset, there are four following classes used in annotations: \textit{Clear}, \textit{Transparent}, \textit{Semi-Transparent}, \textit{Opaque}. The class distribution is heavily imbalanced, where each class contains approximately the following number of pixels: 
\textit{Clear} - 39\% of pixels, \textit{Transparent} - 12\%, \textit{Semi-Transparent} - 19\%, and \textit{Opaque} - 30\%.

%\subsection{Class distribution}
%Our analysis now dive into the distribution of annotations within the dataset. Initially, we compute the percentage of coverage for each type of occlusion present. These percentages are then organized into bins using a 10\% split, and a histogram is generated to depict the distribution visually.

%Upon close examination, we note that the predominant frequency of coverage for most occlusion types falls within the 0-10\% range of the image. However, a subtle decreasing trend in frequency per coverage bin is discernible, except for the \textit{Opaque} class. Notably, there is a noteworthy instance of \textit{Opaque} class, where over 10\% of images exhibit an \textit{Opaque} occlusion covering more than 90\% of the image.

%\begin{table}[ht]
%     \centering
%     \caption{The coverage of soiling classes within the image segmented into bins of 10\%.}
%     \label{table:distribution_of_class_coverage}
%     \begin{tabular}{@{} *5c @{}}    \toprule
%     Image coverage&Clear&Transparent&Semi-Transparent&Opaque\\\midrule
%        0\%-10\%&1768&3447&2667&1820\\
%        10\%-20\%&459&496&564&424\\
%        20\%-30\%&283&296&465&646\\
%        30\%-40\%&304&219&429&395\\
%        40\%-50\%&215&202&253&444\\
%        50\%-60\%&317&134&190&339\\
%        60\%-70\%&357&121&139&366\\
%        70\%-80\%&173&73&129&257\\
%        80\%-90\%&417&1&88&154\\
%        90\%-100\%&707&11&76&155\\\bottomrule
%     \hline
%     \end{tabular}
%\end{table}

\subsection{Data split} \label{data_split}
Even though the original dataset is split into test and train subsets, we don't recommend using the same structure. We identify a data leakage between these two original folders. As was already mentioned, images are taken in sequence, where usually 7 or 8 images are from the same scene in sequence and from the same camera. These 7 or 8 images are split between the train and test folder, which means we would be training the network on the same scene as we are later concluding the test.

As there is a data leak between the Valeo train and test subset, we provided our own data split, which addresses this problem. We collected all images and made split based on sequences. We also reduced the number of test images, so we have a bigger training set totalling 4503 images for training and 497 images for test. It should be noted that when we are referring to the train and test set in the rest of the paper, we are always referring to this new refined split.

\subsection{Class definition revision}
Within the original dataset, we encounter four distinct labels as mentioned in~\cite{Ensemble_based_semi_supervised_soiling_learning} with the following definitions:
\begin{itemize}
    \item \textit{Clear}: indicates no soiling present.
    \item \textit{Semi-transparent} and \textit{Transparent}: are blurry regions where background colors are visible with diminished texture. The difference between the \textit{Semi-transparent} and \textit{Transparent} classes is the ability to recognize the object through the blur.
    \item \textit{Opaque}: represents that in that particular region, it is impossible to see through.
\end{itemize}

However, the differentiation between the \textit{Transparent} and \textit{Semi-transparent} classes is somewhat ambiguous according to this definition.

To address this ambiguity, we propose a refined classification scheme, related to the soiling:
\begin{itemize}
    \item \textit{Clear}: Represents scenarios without any kind of occlusion.
    \item \textit{Transparent}: Encompasses regions where colors may be altered by occlusion or objects may appear blurred, yet lines and structural elements maintain their integrity. Additionally, colors remain consistent but may exhibit tonal variations. Examples could be a small deposit of dust.
    \item \textit{Semi-transparent}: Describes regions where object deformities are evident in contrast to \textit{Transparent} regions. For example, distorted lines or unrecognizable colors may be present, though objects remain visible and distinguishable. Example could be more dust on camera or water drops.
    \item \textit{Opaque}: Signifies regions where visibility is entirely obscured by non-transparent colors, preventing observation of what lies beyond. Example could be water drop that are distorting section of image that much, as its not possible to see through. Another example could be mud or leafs. 
\end{itemize}

\begin{figure}[h]
    \includegraphics[width=\textwidth]{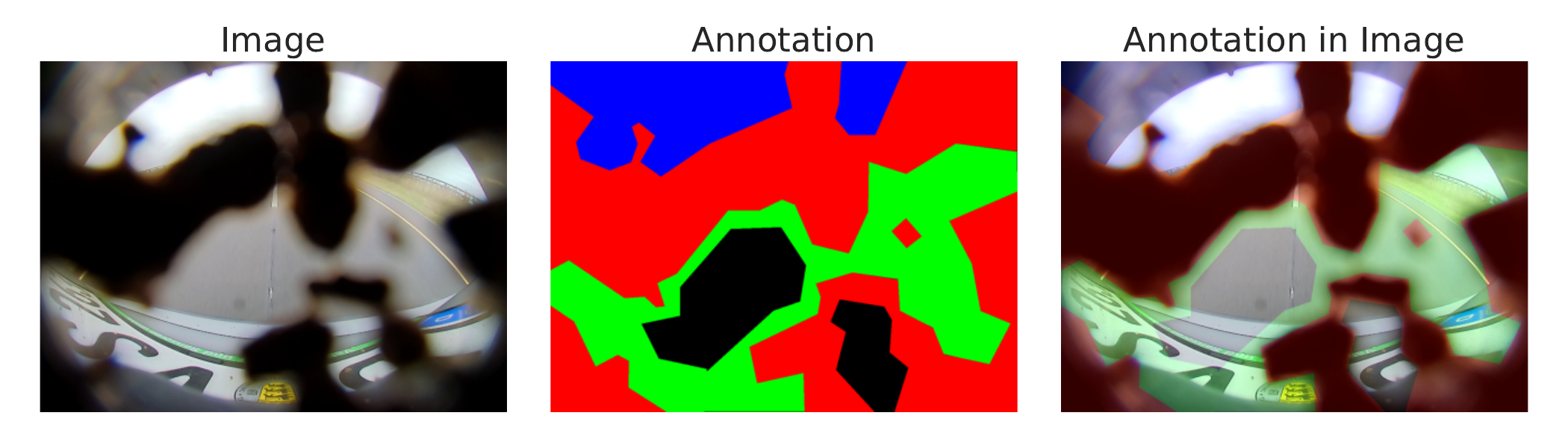}

    \caption{Example of a soiling annotation from the Woodscape dataset. Black is \textit{Clear}, green is \textit{Transparent}, blue is \textit{Semi-transparent}, and red is \textit{Opaque} class}
    \label{fig:example_of_annotation}
\end{figure}

An illustrative example of these annotations can be seen in Fig. \ref{fig:example_of_annotation}

\subsection{Annotation revision}
\begin{figure}[]
    \includegraphics[width=\textwidth]{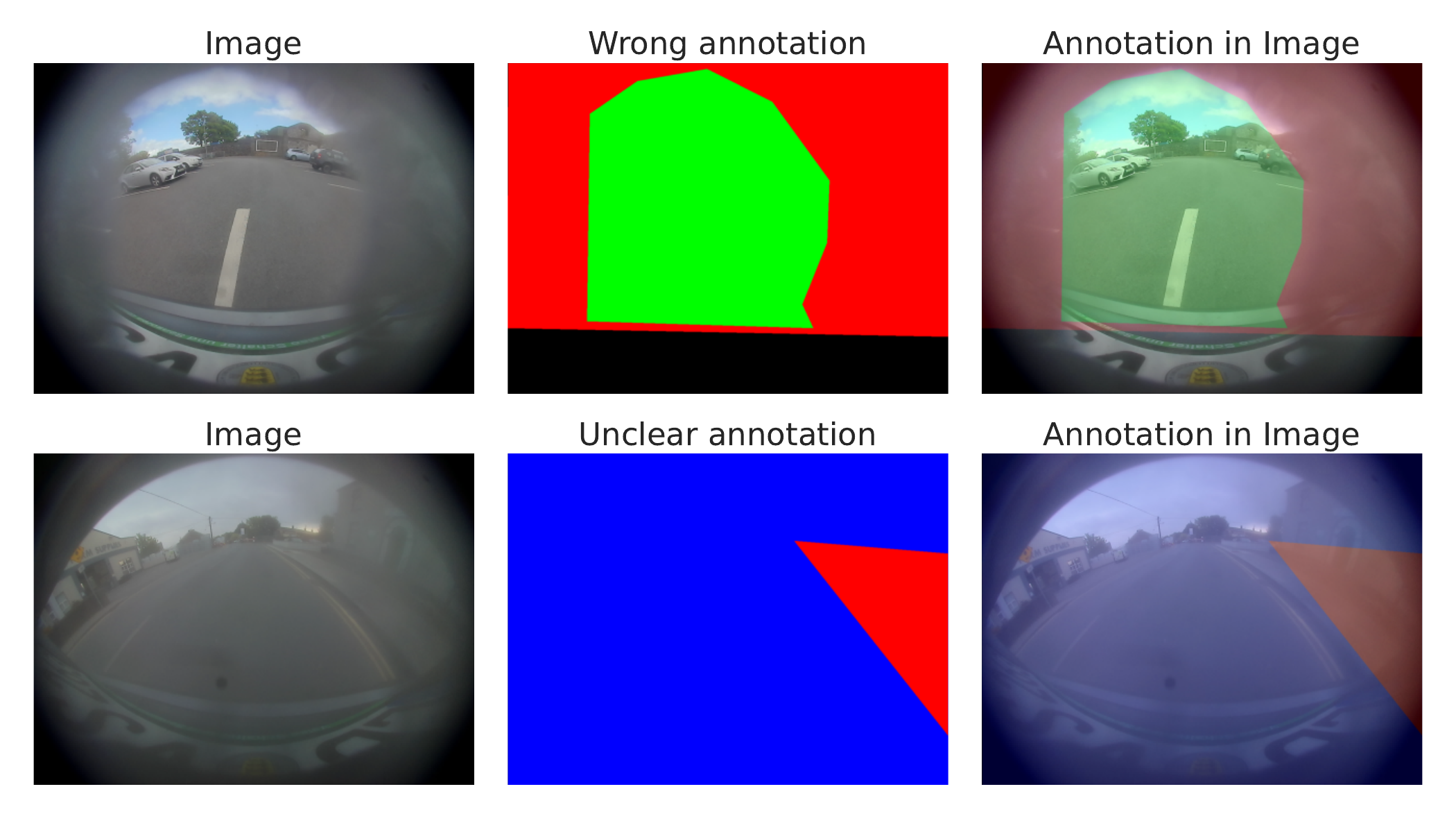}
    \caption{Example illustrating the issues with annotations. The first row exhibits a clear annotation error, while the second row showcases a simplification in class differentiation.}
    \label{fig:mistakes_in_annotations}
\end{figure}

As aforementioned the Woodscape dataset was not flawlessly annotated. The following study~\cite{Ensemble_based_semi_supervised_soiling_learning} points out the potential mismatches in annotations. Additionally, articles such as~\cite{Soilingnet} and~\cite{TiledSoilingNet} address inconsistencies and challenges in distinguishing between classes. Consequently, we had to conduct a manual inspection of the dataset, identifying three key labeling scenarios:

\begin{itemize}
    \item Image is correctly annotated.
    \item Completely erroneous annotations, that could be confusing. An example of such annotations is illustrated in Fig. \ref{fig:mistakes_in_annotations}, depicting a cut-off annotation in the first row, affecting the following 685 images.
    \item Ambiguities in annotations; which are especially notable within the \textit{Transparent} and \textit{Semi-transparent} classes, where the distinction between the two classes is not sharp. The lack of clear demarcation between these classes can potentially impact the outcomes. Such an ambiguity affected additional 519 cases.
    \item Inconsistencies in class definitions, where the same scene may be labeled as \textit{Transparent} in one instance and \textit{Semi-transparent} in another. This inconsistency, often driven by subjectivity, contributed to further training and evaluation challenges. Approximately 1375 images were affected by such inconsistencies.
\end{itemize}

Examples of these annotation challenges are depicted in the Fig. \ref{fig:mistakes_in_annotations}.

% Despite our effort, human subjectivity and difficulty in strict class definition remains a challenge. 

% Addressing the issue mentioned in Section\ref{data_split} with the data leakage and annotation quality, we created baseline data split into training and test sets, consisting of 4503 images for training and 497 images for testing. This testing set will be used for the evaluation of all experiments. All of the images in the testing dataset have been selected without issues mentioned in \textit{Dataset issues}. This means we focus on images with precise and accurate annotations. The rest of the images have been used for the training dataset totalling 4503 images for training. 

To address this problem, we provide three additional training subsets based on the precision of the annotations in them. We denote our original split from Subsection~\ref{data_split} as \textit{Baseline set}. After that, we take the training part of the \textit{Baseline set} and split it into two parts - training set and validation set. Each of these subsets is then manually cleaned by an expert annotator. We opt to let only one expert clean the whole dataset so we prevent subjective differences between two different experts. At the end of this procedure, we get four training and four validation sets in total. The distribution of class-related pixels for these sets can be found in Table~\ref{table:train_test}.

\begin{table}[ht]
     \centering
     \caption{Split of data into train and validation sets, and distribution of classes for each set. Table is sum of class related pixels in millions.}
      \label{table:train_test}
     \begin{tabular}{@{} lccc @{}}    \toprule
     Annotation class & Training set & Validation set \\\midrule
    Clear &  1981 & 378\\ 
    Transparent &   716 & 36\\ 
    Semi-transparent & 1133 & 53\\ 
    Opaque &  1702 & 142\\ \bottomrule 
     \hline
     \end{tabular}
 \end{table}

The presented subsets are defined as follows:
\begin{itemize}
    \item \textit{Baseline set}- comprising all files, including those with errors.
    \item \textit{Correct files} - excluding obvious errors.
    \item \textit{Correct clear files} - eliminating unclear annotations and simplifications in addition to errors.
    \item \textit{Correct clear strict files} - further refining by removing inconsistencies in class assignments.
\end{itemize}

It should be noted that we provide only one original test set for all the subsets. This step ensures a fair comparison between these subsets. The class distribution of the individual subsets can be found in Table~\ref{table:split_of_training_and_val_dataset}.

\begin{table}[ht]
     \centering
     \caption{Split of data into train and val sets, and distribution of classes for each set. Table is sum of class related pixels in millions.}
     \label{table:split_of_training_and_val_dataset}
     \begin{tabular}{@{} lcc |lcc @{}}    \toprule
     \multicolumn{3}{c}{Baseline set} & \multicolumn{3}{c}{Correct files}\\ \hline
     Class & Training & Validation & Class & Training & Validation \\\midrule
    Clear &  1773 & 208 & Clear &  1649 & 177\\ 
    Transparent &   590 & 125 & Transparent &   463 & 87\\ 
    Semi-transparent & 926 & 206 & Semi-transparent & 675 & 181\\ 
    Opaque &  1419 & 282 & Opaque &  1079 & 222\\ \bottomrule 
    \multicolumn{3}{c}{Correct clear set} & \multicolumn{3}{c}{Correct clear strict files}\\ \hline
    Class & Training & Validation & Class & Training & Validation \\\midrule
    Clear &  1500 & 177 & Clear &  889 & 130\\ 
    Transparent &   281 & 69 & Transparent &   121 & 21\\ 
    Semi-transparent & 523 & 134 & Semi-transparent & 154 & 13\\ 
    Opaque &  924 & 145 & Opaque &  375 & 62\\ \bottomrule 
     \hline
     \end{tabular}
 \end{table}

% Vzali jsme tyhle modely jako baseline models 
% Experimental setup
% Smazat dataset z columns
% Napsat na kterem datasetu to je
% Smazat encoder 
% 3 desetinna mista
% V teto kapitole pracujeme s timto dataset
% Od nejmensi po nejvetsi
% Pridat citace do tabulky
% FP Net
% Dat tabulku od nejmensiho resnetu po nejvetsi
% Pridat citace do tabulky

\section{Experiments}
In this chapter, we present a series of experiments. Firstly, we focus on the task of semantic segmentation enriching the previous results from Valeo. Within these experiments, we also study the impact of different training setups, namely the choice of encoder size and choice of loss function. Secondly, we validate the training data and its impact on the segmentation results. As mentioned in Section~\ref{03_dataset_analysis}, we provide four subsets of data addressing the problem with annotations strictness. We compare the performance of the chosen semantic segmentation method and also its training t

If not stated otherwise, we are using the following experimental setup. We start with a learning rate of 0.01 and schedule it with a patience factor of 5 epochs with $\epsilon$ of 0.2. We also use an early stop condition with a patience of 10 epochs and a minimum loss bigger than 0. Learning rate scheduler as well as early stop conditions are related to validation accuracy. The maximum number of epochs is set to 400. We use Resnet18~\cite{he2016deep} as an encoder. Batch size is equal to 5. As an optimizer, we use Adam with $\epsilon$ $1^{-8}$. Finally, we conduct training on CPU AMD Ryzen 7 7840HS having RAM 32GB LPDDR5x and graphic card NVIDIA GeForce RTX 4060 with 8GB VRAM.

\subsection{Networks experiments}
Two notable approaches and their outcomes, as documented by Valeo in~\cite{Soilingnet} and~\cite{TiledSoilingNet} can be used as a baseline for our further experiments. SoilingNet utilizes a ResNet10 encoder and a unique decoder designed for tile-level classification. This setup employs tile dimensions of 64 x 64 pixels. Both models were trained and tested on the complete dataset comprising 76,448 images. However, discrepancies arise in the label definitions used, as only three labels—"Clean," "Transparent," and "Opaque"—are employed, differing from the annotations in the published dataset and subsequent works such as~\cite{TiledSoilingNet}. Moreover, the 64 $\times$ 64 pixels tile size employed by SoilingNet poses challenges for comparison with semantic segmentation methodologies.

In contrast, TiledSoilingNet~\cite{TiledSoilingNet} adopts the same label definitions but operates at a higher classification granularity, namely a tile size of 4 x 4 pixels. This finer granularity aligns better with semantic segmentation, allowing us to compare the results. It should be noted that the results are not directly comparable, because Valeo used the whole datasets in their previous research, whereas in our research we are only able to use its small publicly-available fraction. 

To be able to compare our approaches with the proposed baseline, we employ an accuracy metric despite not being optimal for unbalanced datasets. All our listed results are the results from the test set.

In this study, we utilize the following popular semantic segmentation approaches: Unet \cite{unet}, Unet++ \cite{unetplutplus}, DeepLabV3 \cite{deeplabv3}, DeepLabV3++ \cite{deeplabv3plusplus}, FPNet \cite{fpn}, PSPNet \cite{pspnet}, MaNet \cite{manet}, LinkNet \cite{linknet}, and PAN \cite{pan}.

% We report results obtained from a model trained and tested across all cameras, akin to our approach. There are other results measuring dependency on position, but for our experiments, we use all of the camera positions. Apart from the confusion matrix, we also calculate \textbf{accuracy} which will be the main metric to be compared.

%Our "Baseline" dataset, contains all of the files f training. Where and for testing we use our testing set. As loss function we use same as in~\cite{TiledSoilingNet}, which is RMSE.

As can be seen in Table~\ref{table:Networks_comparison} the best performance was achieved by FPNet with an accuracy of 94.0\%. All the tested semantic segmentation methods surpass the original tile-level classification approach.

\begin{table}[ht]
     \centering
     \caption{Accuracy of TileSoilingNet providing tiled classification based on 4x4 pixel matrix compared to semantic segmentation networks using same loss and similar size of encoder as aforementioned.}
     \label{table:Networks_comparison}
     \begin{tabular}{@{} lc @{}}    \toprule
     Architecture & $\uparrow$ Accuracy \\\midrule
     TiledSoilingNet\cite{TiledSoilingNet}& 0.874\\\midrule
     Deeplabv3\cite{deeplabv3plusplus}& 0.912\\
     Unet++\cite{unetplutplus}&  0.925\\
     Deeplabv3+\cite{deeplabv3}&  0.927\\
     Unet\cite{unet}&  0.929\\ 
     Pan\cite{pan}&  0.930\\ 
     MANet\cite{manet}&  0.934\\ 
     PSPNet\cite{pspnet}&  0.934\\ 
     LinkNet\cite{linknet}& 0.936\\ 
     FPNet\cite{fpn}& \textbf{0.940}\\\bottomrule
     \hline
     \end{tabular}
 \end{table}

To further explore possibilities of semantic segmentation approaches we take the best approach - FPNet - and test it with encoders of five different sizes. The results can be found in Table~\ref{table:encoder_acc}. It can seen that the difference between encoders is mostly negligible. We believe that this phenomenon is caused by the relatively small size of the training dataset. That means that bigger encoders can't utilize their better computational capacity.

\begin{table}[ht]
     \centering
     \caption{Experiment displaying the correlation between encoder size and results on Woodscape dataset using FPNet.}
     \label{table:encoder_acc}
     \begin{tabular}{@{} lc @{}}    \toprule
     Encoder &$\uparrow$Accuracy \\\midrule
        Resnet18&  0.940\\
        Resnet34&  0.942\\  
        \textbf{Resnet50}&  \textbf{0.945}\\ 
        Resnet101& 0.935\\ 
        Resnet152& 0.939\\\bottomrule
     \hline
     \end{tabular}
 \end{table}
 
As mentioned in~\cite{TiledSoilingNet}, another factor that can affect the result is the loss function. In the original paper, the authors claim that the best loss for this task was RMSE. To verify this conclusion, we conducted an additional ablation study using different losses as depicted in Table~\ref{table:loss_correlation}. In our experimental setup, the best results are reached while training with Cross-Entropy loss, nevertheless, the differences between different setups are not statistically significant again.

\begin{table}[ht]
     \centering
     \caption{Experiment displaying correlation of loss and accuracy using FPNet with different loss functions.}
     \label{table:loss_correlation}
     \begin{tabular}{@{} lc @{}}    \toprule
     Loss function & $\uparrow$Accuracy \\\midrule
    Focal &  0.939\\ 
    Dice &   0.930\\ 
    Cross-Entropy & \textbf{0.944}\\ 
    RMSE &  0.940\\ \bottomrule 
     \hline
     \end{tabular}
 \end{table}

\subsection{Training sets experiments}
In this subsection, we focus on different data subsets. As mentioned earlier, we create four different data subsets based on the annotation's strictness. We choose the five best networks from the previous subsection to compare their performance further. We train all of the networks with a Categorical Cross-entropy loss function. The rest of the training hyperparameters remain unchanged.

\begin{figure}[]
    \includegraphics[width=\textwidth]{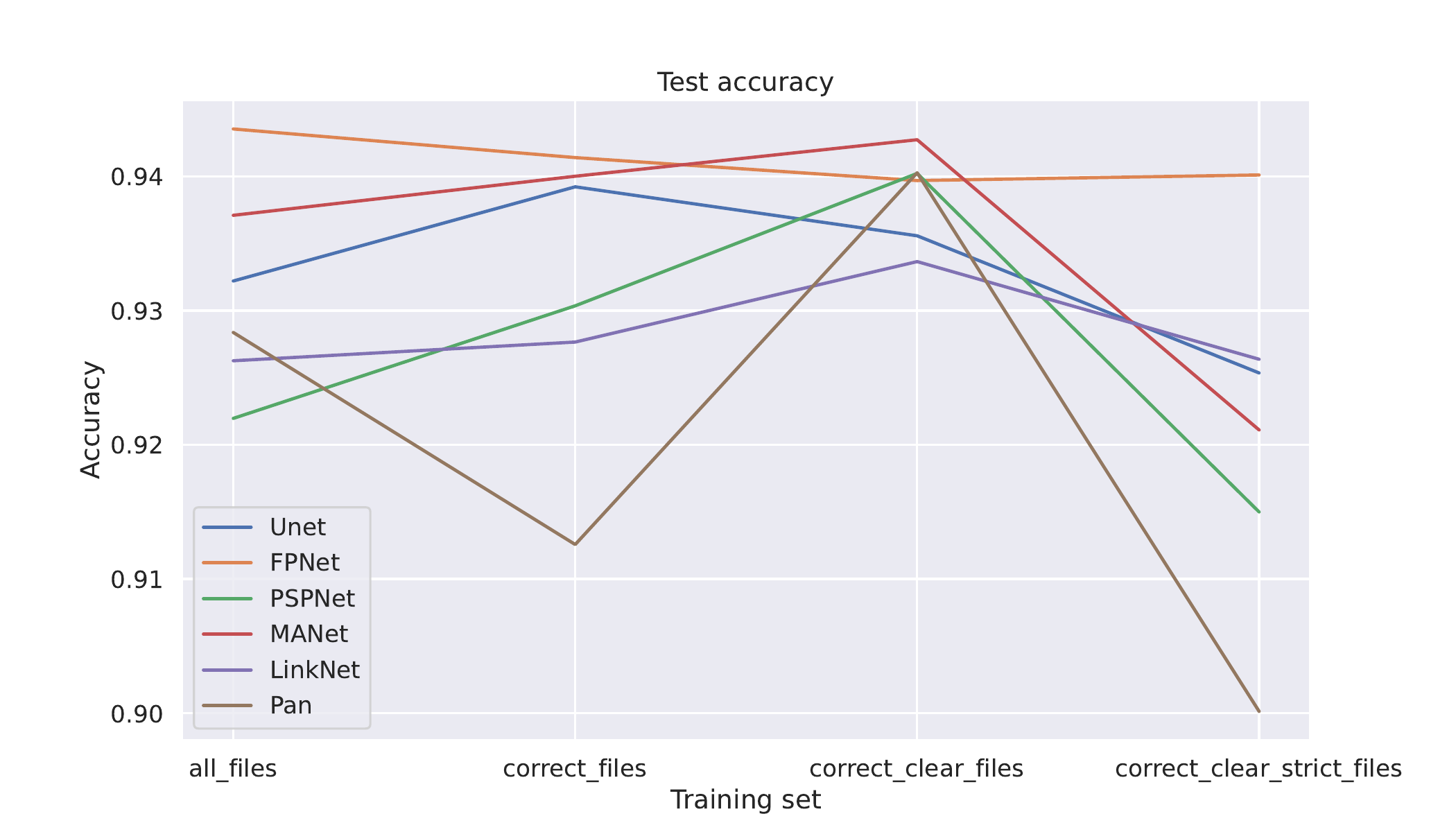}
    \caption{Measured test accuracies across all networks and different training sets.}
    \label{fig:test_accuracies}
\end{figure}

\begin{figure}[]
    \includegraphics[width=\textwidth]{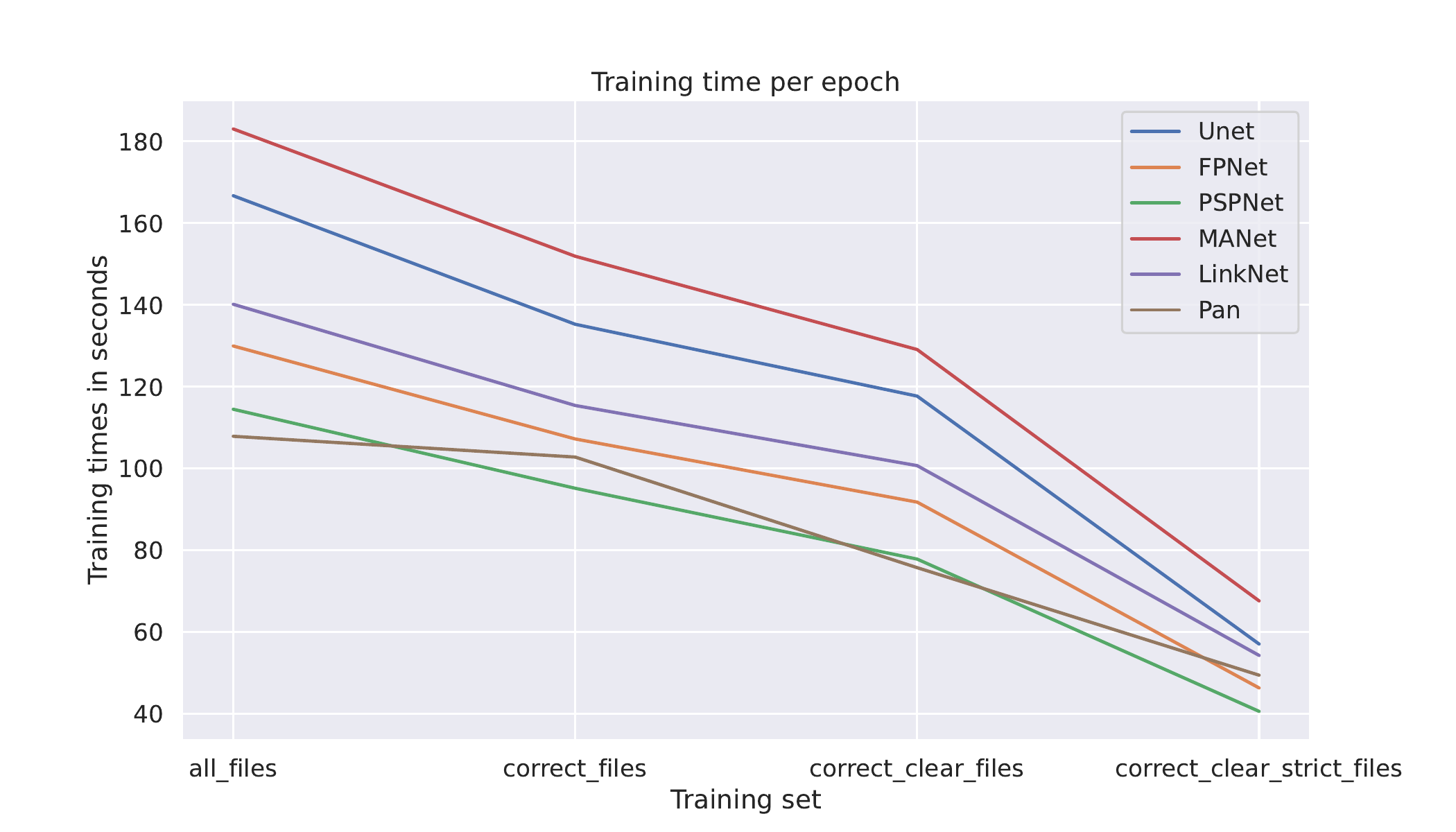}
    \caption{Measured training time per epoch across all networks and different training sets.}
    \label{fig:test_training_time}
\end{figure}

A comparison of training accuracies and training times per one epoch can be found in Fig.~\ref{fig:test_accuracies} and Fig.~\ref{fig:test_training_time}, respectively. 

The performance for different subsets is comparable except the last subset - \textit{Correct clear strict files} where the significant drop in performance can be observed for the most tested methods. We argue this is expected behavior due to the large difference between the number of files in the last subsets. For the best trade-off between the training time and the model's accuracy, the usage of \textit{Correct clear files} subset is recommended.

\section{Conclusion}
In this paper, we have presented a deep analysis of the Woodscape dataset that reveals several issues. The first issue - data leakage - was addressed by proposing a new data split. An additional analysis revealed vague class definitions, which can cause trouble while annotating new data. We extended these original class definitions to make them more precise and, therefore, prevent possible problems in the future. Lastly, we revealed problems with already existing annotations, and based on the manual inspection of the data we created four new subsets of the training data.

In the second part of the paper, we experimented with chosen semantic segmentation methods on the problem of soiling detection. These experiments show the superiority of the semantic segmentation approach while all of the tested methods surpass the previous research, which handles the soiling detection as tile-level classification. Moreover, we provide an ablation study on the topic of the choice of encoder size and also the choice of the loss function.

In our future research, we would like to focus on additional ablation studies of training hyperparameters and also test transformer-based approaches. Also obtaining additional training data can be beneficial for the whole scientific community.

%In this paper we have presented issues with Woodscape dataset starting from data leak from data split. Then we presented that there is unclear class definition in the dataset, therefore we came with new and more precise definition. We review annotations of dataset and found errors in annotations. We also found some discrepancies in annotations between classes and some subjectivity between scenes. From these issues, we have created 4 sets of data for training and test, which could be use.

%We review and compare results of TiledSoilingNet to semantic segmentation networks like the first one. It shown, that semantic segmentation is more accurate, than TileSoilingNet. Also we experimented that better loss for this dataset and task would be focal loss rather than RMSE.

%Moreover we experimented with size of Resnet encoder, where we prove that it has impact with increasing size, but training time is much bigger and gain is smaller.

%During tests with different test, its been shown that networks are able to generalize and absorbs some errors, but with use of our sets, that are significantly smaller thus training is faster, there can be reached same results as with complete dataset.

%We belive that this could lead into further development of solutions that would work under adheres conditions. 
\section*{Acknowledgements}
The work has been supported by the grant of the University of West Bohemia, project No. SGS-2022-017 and by ARRK Engineering GmbH. Computational resources were provided by the e-INFRA CZ project (ID:90254), supported by the Ministry of Education, Youth and Sports of the Czech Republic.

\bibliographystyle{splncs04}
\bibliography{bibliography}

\end{document}